\documentclass[twocolumn,times,final,authoryear]{elsarticle}
\usepackage[latin9]{inputenc}
\usepackage{array}
\usepackage{calc}
\usepackage{url}
\usepackage{multirow}
\usepackage{amsmath}
\usepackage{amssymb}
\usepackage{graphicx}

\makeatletter

\providecommand{\tabularnewline}{\\}

\usepackage{prletters}
\usepackage{framed}
\usepackage{multirow}

\usepackage{latexsym}

\usepackage{url}
\usepackage{xcolor}
\definecolor{newcolor}{rgb}{.8,.349,.1}

\journal{Pattern Recognition Letters}

\@ifundefined{showcaptionsetup}{}{%
 \PassOptionsToPackage{caption=false}{subfig}}
\usepackage{subfig}
\makeatother

\begin{document}
\thispagestyle{empty}
 
\begin{frontmatter}{}

\title{Deep Learning with $t$-Exponential Bayesian Kitchen Sinks}

\author[1]{Harris \snm{Partaourides}}

\ead{c.partaourides@cut.ac.cy}

\author[1]{Sotirios \snm{Chatzis}\corref{cor1}}

\ead{sotirios.chatzis@cut.ac.cy}

\address[1]{Department of Electrical Eng., Computer Eng., and Informatics\\
Cyprus University of Technology, Limassol 3036, Cyprus}

\received{1 May 2013} \finalform{10 May 2013} \accepted{13
May 2013} \availableonline{15 May 2013} \communicated{S. Sarkar}
\begin{abstract}
Bayesian learning has been recently considered as an effective means
of accounting for uncertainty in trained deep network parameters.
This is of crucial importance when dealing with small or sparse training
datasets. On the other hand, shallow models that compute weighted
sums of their inputs, after passing them through a bank of arbitrary
randomized nonlinearities, have been recently shown to enjoy good
test error bounds that depend on the number of nonlinearities. Inspired
from these advances, in this paper we examine novel deep network architectures,
where each layer comprises a bank of arbitrary nonlinearities, linearly
combined using multiple alternative sets of weights. We effect model
training by means of approximate inference based on a $t$-divergence
measure; this generalizes the Kullback-Leibler divergence in the context
of the $t$-exponential family of distributions. We adopt the $t$-exponential
family since it can more flexibly accommodate real-world data, that
entail outliers and distributions with fat tails, compared to conventional
Gaussian model assumptions. We extensively evaluate our approach using
several challenging benchmarks, and provide comparative results to
related state-of-the-art techniques.
\end{abstract}
\begin{keyword}
Random kitchen sinks; Student's-$t$ Distribution; $t$-Divergence;
variational Bayes.
\end{keyword}

\end{frontmatter}{}

\section{Introduction}

Deep neural networks (DNNs) have experienced a resurgence of interest;
this is due to recent breakthroughs in the field that offer unprecedented
empirical results in several challenging real-world tasks, such as
image and video understanding \citep{he}, natural language understanding
and generation \citep{sutskever}, and game playing \citep{silver}.
Most DNN models are trained by means of some variant of the backpropagation
(BP) algorithm. However, despite all these successes, BP suffers from
the major shortcoming of being able to obtain only point-estimates
of the trained networks. This fact results in the trained networks
generating predictions that do not account for uncertainty, e.g. due
to the limited or sparse nature of the available training data. 

A potential solution towards the amelioration of these issues consists
in treating some network components under the Bayesian inference rationale,
instead of stochastic optimization \citep{neal}. Indeed, Bayesian
inference principles have been recently met with great success in
the context of DNN regularization, e.g. \citep{dropout}. Specifically,
in this paper we are interested in inference of the network \emph{feature
functions}. In the literature, this is effected by considering them
as stochastic latent variables imposed some mathematically convenient
Gaussian process prior \citep{vb-weights}. On this basis, one proceeds
to infer the corresponding posteriors, based on the available training
data. To the latter end, and with the goal of combining accuracy with
computational efficiency, expectation-propagation \citep{ep}, mean-field
\citep{vb-weights,tnnls15}, and probabilistic backpropagation \citep{probbp}
have been used.

One of the main driving forces behind the unparalleled data modeling
and predictive performance of modern DNNs is their capability of effectively
learning to extract informative, high-level, hierarchical representations
of observed data with latent structure \citep{lecun}. Nevertheless,
DNNs are not the only class of models that entail this sort of functionality.
Indeed, major advances in machine learning have long been dominated
by the development of shallow architectures that compute weighted
sums of feature functions; the latter generate nonlinear representations
of their input data, which can be determined under a multitude of
alternative rationales. For instance, models that belong to the family
of support vector machines (SVMs) \citep{svm-book} essentially compute
weighted sums of positive definite kernels; boosting algorithms, such
as AdaBoost \citep{adaboost}, compute weighted sums of weak learners,
such as decision trees. In all cases, both the feature functions as
well as the associated weights are learned under an empirical risk
minimization rationale; for instance, the hinge loss is used in the
context of SVMs, while AdaBoost utilizes the exponential loss. 

In the same vein, the machine learning community has recently examined
a bold, yet quite promising possibility: postulating weighted sums
of \emph{random kitchen sinks }(RKS)\emph{ }\citep{kitchen}\emph{.
}The main rationale of this family of approaches essentially consists
in randomly drawing the employed feature functions (nonlinearities),
and limiting model training to the associated (scalar) weights. Specifically,
the (entailed parameters of the) postulated nonlinearities are randomly
sampled from an appropriate density, which is a priori determined
by the practitioners according to some assumptions \citep{fastfood,alacarte}.
As it has been shown, both through theoretical analysis as well as
some empirical evidence, such a modeling approach does not yield much
inferior performance for a trained classifier compared to the mainstream
approach of optimally selecting the employed nonlinearities. In addition,
predictive performance is shown to increase with the size of the employed
bank of randomly drawn nonlinearities. 

Inspired from these advances, this paper introduces a fresh regard
towards DNNs: We formulate each DNN layer as a bank of random nonlinearities,
which are linearly combined in multiple alternative fashions. This
way, the postulated models eventually yield a hierarchical cascade
of informative representations of their multivariate observation inputs,
that can be used to effectively drive a penultimate regression or
classification layer. At each layer, the employed bank of random nonlinearities
is sampled from an appropriate postulated density, in a vein similar
to RKS. However, in order to alleviate the burden of having to manually
design these densities, we elect to infer them in a Bayesian sense.
In addition, we elect not to obtain point-estimates of the weights
used for combining the drawn nonlinearities. On the contrary, we perform
Bayesian inference over them; this way, we allow for strong model
regularization, by better accounting for model uncertainty, especially
when training data availability is limited or sparse.

As already discussed, Bayesian inference for DNN-type models can be
performed under various alternative paradigms. Here, we resort to
variational inference ideas, which consist in searching for a proxy
in an analytically solvable distribution family that approximates
the true underlying posterior distribution. To measure the closeness
between the true and the approximate posterior, the relative entropy
between these two distributions is used. Specifically, under the typical
Gaussian assumption, one can use the Shannon-Boltzmann-Gibbs (SBG)
entropy, whereby the relative entropy yields the well known Kullback-Leibler
(KL) divergence \citep{entropy}. However, real-world phenomena tend
to entail densities with heavier tails than the simplistic Gaussian
assumption.

To account for these facts, in this work we exploit the $t$-exponential
family\footnote{Also referred to as the $q$-exponential family or the Tsallis distribution.},
which was first proposed by Tsallis and co-workers \citep{tsalis,tsalis2,tsalis3},
and constitutes a special case of the more general $\phi$-exponential
family \citep{naudts,naudts2,naudts3}. Of specific practical interest
to us is the Students'-$t$ density; this is a bell-shaped distribution
with heavier tails and one more parameter (degrees of freedom - DOF)
compared to the normal distribution, and tends to a normal distribution
for large DOF values \citep{key-19}. This way, it offers a solution
to the problem of providing protection against outliers in multidimensional
variables \citep{key-19,vbsmm,vbsmmBishop}, which is a very difficult
problem that increases in difficulty with the dimension of the variables
\citep{key-30}. On top of these merits, the $t$-exponential family
also gives rise to a new $t$-divergence measure; this can be used
for performing approximate inference in a fashion that better accommodates
heavy-tailed densities (compared to standard KL-based solutions) \citep{tVB}. 

To summarize, we formulate a hierarchical (multilayer) model, each
layer of which comprises a bank of random feature functions (nonlinearities).
Each nonlinearity is presented with the layer's input, and generates
scalar outputs. These outputs are linearly combined by using multiple
alternative sets of mixing weights, to produce the (multivariate)
layer's output. At each layer, the postulated nonlinearities are drawn
from an appropriate (posterior) density, which is inferred from the
data (as opposed to the requirement of conventional RKS that the practitioners
manually specify this distribution). Indeed, both the posterior density
of the nonlinearities, as well as the posterior over the mixing weights,
are inferred in an approximate Bayesian fashion. This renders our
model more capable of accommodating limited or sparse training data,
while retaining its computational scalability. In addition, in order
to allow for our model to account for heavy tails, we postulate that
the sought densities belong to the $t$-exponential family, specifically
they constitute (multivariate) Student's-$t$ densities. On this basis,
we conduct approximate inference by optimizing a $t$-divergence functional,
that better leverages the advantages of the $t$-exponential family.
We dub our proposed approach the Deep $t$-Exponential Bayesian Kitchen
Sinks (D$t$BKS) model.

The remainder of this paper is organized as follows: In the following
Section, we provide a brief overview of the methodological background
of our approach. Subsequently, we introduce our approach, and derive
its training and inference algorithms. Then, we perform the experimental
evaluation of our approach, and illustrate its merits over the current
state-of-the-art. Finally, in the concluding Section of this paper,
we summarize our contribution and discuss our results.

\section{Methodological Background}

\subsection{Weighted Sums of Random Kitchen Sinks}

Consider the problem of fitting a function $f:\mathbb{R}^{\delta}\rightarrow\mathcal{Y}$
to a training dataset comprising $N$ input-output pairs $\{\boldsymbol{x}_{n},y_{n}\}_{n=1}^{N}$,
drawn i.i.d. from some unknown distribution $P(\boldsymbol{x},y)$,
with $\boldsymbol{x}_{n}\in\mathbb{R}^{\delta}$ and $y_{n}\in\mathcal{Y}$.
In essence, this fitting problem consists in finding a function $f$
that minimizes the empirical risk on the training data
\begin{equation}
R[f]=\frac{1}{N}\sum_{n=1}^{N}c\left(f(\boldsymbol{x}_{n}),y_{n}\right)
\end{equation}
 where the cost function $c(f(\boldsymbol{x}),y)$ defines the penalty
we impose on the deviation between the prediction $f(\boldsymbol{x})$
and the actual value $y$.

The main underlying idea of data modeling under the weighted sums
of RKS rationale consists in postulating functions of the form
\begin{equation}
f(\boldsymbol{x})=\sum_{s=1}^{S}\alpha_{s}\xi(\boldsymbol{x};\boldsymbol{\omega}_{s})
\end{equation}
where the $\{\alpha_{s}\}_{s=1}^{S}$ are mixing weights, while the
nonlinear feature functions $\xi$ are parameterized by some vector
$\boldsymbol{\omega}\in\Omega$, and are bounded s.t.: $|\xi(\boldsymbol{x};\boldsymbol{\omega}_{s})|\leq1$.
Specifically, the vectors $\boldsymbol{\omega}_{s}$ are samples drawn
from an appropriate probability distribution $p(\boldsymbol{\omega})$
with support in $\Omega$, whence we have $\xi:\mathbb{R}^{\delta}\times\Omega\rightarrow[-1,1]$. 

As an outcome of this construction, the fitted function $f(\boldsymbol{x})$
can be essentially viewed as \emph{a weighted sum of a bank of random
nonlinearities}, $\xi$, drawn by appropriately sampling from a selected
probability distribution $p(\boldsymbol{\omega})$. On this basis,
the fitting procedure reduces to\emph{ selecting the weight values}
$\{\alpha_{s}\}_{s=1}^{S}$, such that we minimize the empirical risk
(1); typically, a quadratic loss function is employed to this end.
As shown in \citep{kitchen}, for $S\rightarrow\infty$, weighted
sums of RKS yield predictive models whose true risk is near the lowest
true risk attainable by an infinite-dimensional class of functions
with optimally selected parameter sets, $\boldsymbol{\omega}$.

Weighted sums of RKS give rise to a modeling paradigm with quite appealing
properties: It allows for seamlessly and efficiently employing arbitrarily
complex feature functions $\xi$, since model fitting is limited to
the mixing weights; this way, we can easily experiment with feature
functions that do not admit simple fitting procedures. On the other
hand, RKS require one to (manually) design appropriate distributions
$p(\boldsymbol{\omega})$ to draw the employed nonlinearities from;
in real-world data modeling tasks, this might prove quite challenging
a task. 

\subsection{The Student's-$t$ Distribution}

The adoption of the multivariate Student's-$t$ distribution provides
a way to broaden the Gaussian distribution for potential outliers.
The probability density function (pdf) of a Student's-$t$ distribution
with mean vector $\boldsymbol{\mu}$, covariance matrix $\boldsymbol{\Sigma}$,
and $\nu>0$ degrees of freedom is \citep{liu}
\begin{equation}
t(\boldsymbol{y}_{t};\boldsymbol{\mu},\boldsymbol{\Sigma},\nu)=\frac{\Gamma\left(\frac{\nu+\delta}{2}\right)|\boldsymbol{\Sigma}|^{-1/2}(\pi\nu)^{-\delta/2}}{\Gamma(\nu/2)\{1+d(\boldsymbol{y}_{t},\boldsymbol{\mu};\boldsymbol{\Sigma})/\nu\}^{(\nu+\delta)/2}}
\end{equation}
where $\delta$ is the dimensionality of the observations $\boldsymbol{y}_{t},$
$d(\boldsymbol{y}_{t},\boldsymbol{\mu};\boldsymbol{\Sigma})$ is the
squared Mahalanobis distance between $\boldsymbol{y}_{t},\boldsymbol{\mu}$
with covariance matrix $\boldsymbol{\Sigma}$
\begin{equation}
d(\boldsymbol{y}_{t},\boldsymbol{\mu};\boldsymbol{\Sigma})=(\boldsymbol{y}_{t}-\boldsymbol{\mu})^{T}\boldsymbol{\Sigma}^{-1}(\boldsymbol{y}_{t}-\boldsymbol{\mu})
\end{equation}
and $\Gamma(s)$ is the Gamma function, $\Gamma(s)=\int_{0}^{\infty}e^{-t}z^{s-1}dz$. 

A graphical illustration of the univariate Student's-$t$ distribution,
with $\boldsymbol{\mu}$, $\boldsymbol{\Sigma}$ fixed, and for various
values of the degrees of freedom $\nu$, is provided in Fig. 1. As
we observe, as $\nu\rightarrow\infty$, the Student's-$t$ distribution
tends to a Gaussian with the same $\boldsymbol{\mu}$ and $\boldsymbol{\Sigma}$.
On the contrary, as $\nu$ tends to zero, the tails of the distribution
become longer, thus allowing for a better handling of potential outliers,
without affecting the mean or the covariance of the distribution \citep{vbshmm,shmm}.

\begin{figure}
\begin{centering}
\includegraphics[scale=0.51]{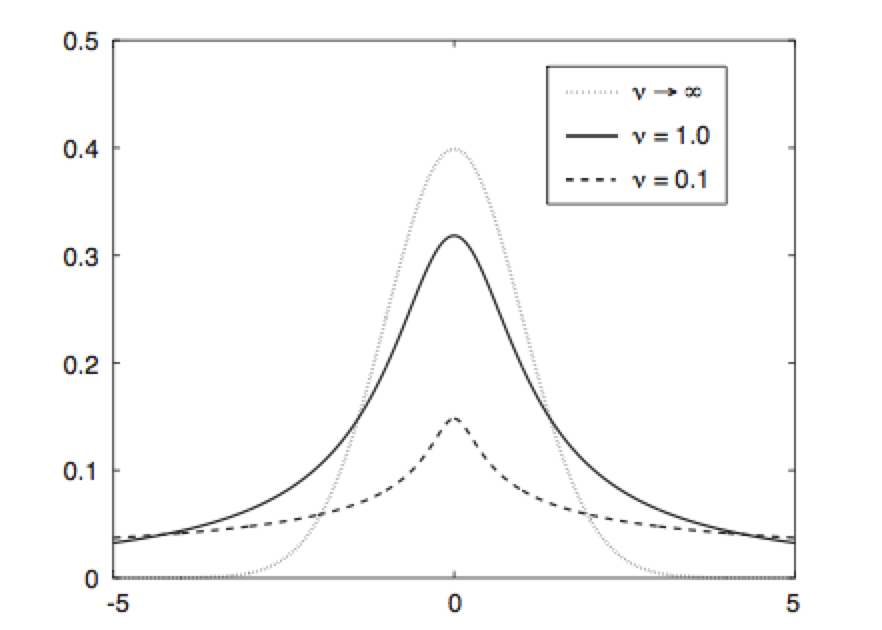} 
\par\end{centering}
\caption{Univariate Student's-$t$ distribution $t(\boldsymbol{y}_{t};\boldsymbol{\mu},\boldsymbol{\Sigma},\nu)$,
with $\boldsymbol{\mu}$, $\boldsymbol{\Sigma}$ fixed, for various
values of $\nu$ \citep{vbsmm2}. }
\end{figure}

\subsection{The $t$-Divergence}

The $t$-divergence was introduced in \citep{tVB} as follows:\\
\textbf{Definition 1. }The $t$-divergence between two distributions,
$q(\boldsymbol{h})$ and $p(\boldsymbol{h})$, is defined as
\begin{equation}
D_{t}(q||p)=\int\tilde{q}(\boldsymbol{h})\mathrm{log}_{t}q(\boldsymbol{h})\mathrm{d}\boldsymbol{h}-\tilde{q}(\boldsymbol{h})\mathrm{log}_{t}p(\boldsymbol{h})\mathrm{d}\boldsymbol{h}\label{eq:td}
\end{equation}
 where $\tilde{q}(\boldsymbol{h})$ is called the escort distribution
of $q(\boldsymbol{h})$, and is given by
\begin{equation}
\tilde{q}(\boldsymbol{h})=\frac{q(\boldsymbol{h})^{t}}{\int q(\boldsymbol{h})^{t}\mathrm{d}\boldsymbol{h}},\;t\in\mathbb{R}
\end{equation}
Importantly, the divergence $D_{t}(q||p)$ preserves the following
two properties:
\begin{itemize}
\item $D_{t}(q||p)\geq0,\;\forall q,p$. The equality holds only for $q=p$.
\item $D_{t}(q||p)\neq D_{t}(p||q)$. 
\end{itemize}
As discussed in \citet{tVB}, by leveraging the above definition of
the $t$-divergence, $D_{t}(q||p)$, one can establish an advanced
approximate inference framework, much more appropriate for fitting
heavy-tailed densities. We exploit these benefits in developing the
training algorithm of the proposed D$t$BKS model, as we shall explain
in the following Section. 

\section{Proposed Approach}

\subsection{Model Formulation}

Let us consider a D$t$BKS model with input variables $\boldsymbol{x}\in\mathbb{R}^{\delta}$
and output (predictable) variables $\boldsymbol{y}$, comprising $L$
layers. In Fig. 2, we provide a graphical illustration of the proposed
configuration of one D$t$BKS model layer. Each layer, $l$, is presented
with an input vector $\boldsymbol{h}^{l-1}$, generated from the preceding
layer; at the first layer, this vector is the model input, $\boldsymbol{h}^{0}\triangleq\boldsymbol{x}$.
This is fed into a bank comprising $S$ randomly drawn feature functions,
$\{\xi_{s}^{l}(\boldsymbol{h}^{l-1})\}_{s=1}^{S}$. Specifically,
these functions are nonlinearities parameterized by some random vector
$\boldsymbol{\omega}$; i.e., $\xi_{s}^{l}(\boldsymbol{h}^{l-1})=\xi(\boldsymbol{h}^{l-1};\boldsymbol{\omega}_{s}^{l})$.
The used samples $\boldsymbol{\omega}_{s}^{l}$ are drawn from an
appropriate density, which is inferred in a Bayesian sense, as we
shall explain next. 

This way, we eventually obtain a set of $S$ univariate signals, which
we linearly combine to obtain the layer's output. Specifically, we
elect to (linearly) combine them in multiple alternative ways, with
the goal of obtaining a potent, multidimensional, high-level representation
of the original observations. These alternative combinations are computed
via a mixing weight matrix, $\boldsymbol{W}^{l}\in\mathbb{R}^{\eta\times S}$,
where $\eta$ is the desired dimensionality of the layer's output
(i.e., the postulated number of alternative linear combinations). 

Eventually, the output vectors at the layers $l\in\{1,\dots,L-1\}$
yield
\begin{equation}
\boldsymbol{h}^{l}=\boldsymbol{W}^{l}[\xi(\boldsymbol{h}^{l-1};\boldsymbol{\omega}_{s}^{l})]_{s=1}^{S}\in\mathbb{R}^{\eta}\label{eq:def-1}
\end{equation}
where $[\chi_{s}]_{s=1}^{S}$ denotes the vector-concatenation of
the set $\{\chi_{s}\}_{s=1}^{S}$. Turning to the penultimate layer
of the model, we consider that the output variables are imposed an
appropriate conditional likelihood function, the form of which depends
on the type of the addressed task. Specifically, in the case of regression
tasks, we postulate a multivariate Gaussian of the form
\begin{equation}
p(\boldsymbol{y}|\boldsymbol{x})=\mathcal{N}\left(\boldsymbol{y}\bigg|\boldsymbol{W}^{L}[\xi(\boldsymbol{h}^{L-1};\boldsymbol{\omega}_{s}^{L})]_{s=1}^{S},\sigma_{y}^{2}\boldsymbol{I}\right)\label{eq:def-2}
\end{equation}
 where $\sigma_{y}^{2}$ is the noise variance. On the other hand,
in case of classification tasks, we assume
\begin{equation}
p(\boldsymbol{y}|\boldsymbol{x})=\mathrm{Softmax}\left(\boldsymbol{y}\bigg|\boldsymbol{W}^{L}[\xi(\boldsymbol{h}^{L-1};\boldsymbol{\omega}_{s}^{L})]_{s=1}^{S}\right)\label{eq:def-3}
\end{equation}

This concludes the definition of our model. For brevity, we shall
omit the layer indices, $l$, in the remainder of this paper, wherever
applicable. As we observe from Eqs. (7)-(9), at each layer of the
proposed model:\\
(i) We postulate a random feature function $\xi(\cdot)$ that generates
scalar outputs. This is effected by taking the inner product of the
layer's input with a parameter vector $\boldsymbol{\omega}$. \\
(ii) We draw many different samples of $\xi(\cdot)$. This is effected
by drawing multiple samples from the parameter vector $\boldsymbol{\omega}$,
which is essentially considered as a random variable.\\
(iii) We optimally combine the drawn samples of $\xi(\cdot)$ in a
linear fashion, by using the trainable matrix $\boldsymbol{W}$. That
is, we compute multiple alternative weighted averages of the drawn
samples of $\xi(\cdot)$, whereby the used sets of weights (stored
in $\boldsymbol{W}$) are trainable parameters. 

Thus, our model constitutes a generalization of random kitchen sinks,
whereby the main rationale consists in drawing multiple random samples
of a fundamental unit. A key difference between the existing literature,
e.g. \citet{alacarte,fastfood,kitchen}, and our approach consists
in the fact that we opt for inferring the distribution we draw samples
from, as opposed to using some fixed selection. In addition, we learn
how to optimally combine these samples, and do it in multiple alternative
ways, so as to generate a multidimensional layer output. Each dimension
of the generated output corresponds to a different way of combining
the drawn samples of the basic unit $\xi(\cdot)$. This is in contrast
to the existing literature on random kitchen sinks, where only shallow
architectures with scalar outputs have been considered \citep{alacarte,fastfood,kitchen},
whereby the drawn samples are combined in one single way.

\begin{figure}
\begin{centering}
\includegraphics[scale=0.17]{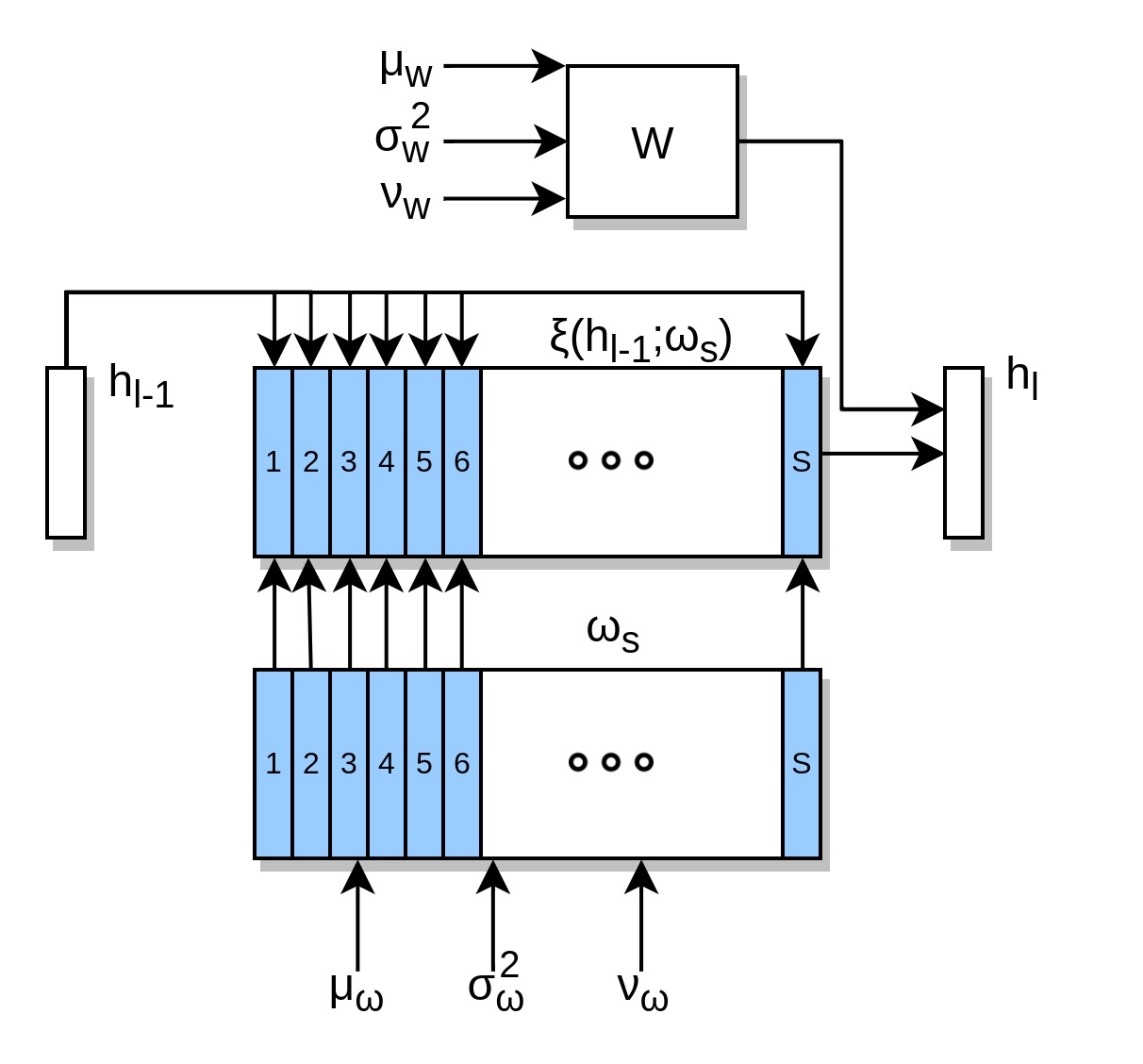} 
\par\end{centering}
\caption{Graphical illustration of the configuration of one D$t$BKS model
layer. }
\end{figure}

\subsection{Model Training}

To perform model training, we opt for a full Bayesian treatment. This
is in contrast to most deep learning approaches, which rely on frequentist
treatments that yield point-estimates of the model parameters. Indeed,
frequentist methods consider model parameters to be fixed, and the
training data to be some random sample from an infinite number of
existing but unobserved data points. On the contrary, our Bayesian
treatment is designed under the assumption of dealing with fixed scarce
data and parameters that are random because they are unknowns. This
way, by imposing a suitable prior over model parameters and obtaining
a corresponding posterior distribution based on the given observed
data, our full Bayesian treatment allows for better including uncertainty
in learning the model; this is important when dealing with limited
or sparse data.

To allow for inferring the distribution that the employed feature
functions, $\xi$, must be drawn from, we first consider that the
vectors $\boldsymbol{\omega}$ that parameterize them are Student's-$t$
distributed latent variables. We employ the same assumption for the
weight matrices, $\boldsymbol{W}$, which we also want to infer in
a Bayesian fashion, so as to account for model uncertainty. Specifically,
we start by imposing a simple, zero-mean Student's-$t$ prior distribution
over them, with tied degrees of freedom, at each model layer: 
\begin{equation}
p(\boldsymbol{\omega})=t(\boldsymbol{\omega}|\boldsymbol{0},\boldsymbol{I},\nu)\label{eq:prior1}
\end{equation}
\begin{equation}
p(\boldsymbol{W})=t(\mathrm{vec}(\boldsymbol{W})|\boldsymbol{0},\boldsymbol{I},\nu)\label{eq:prior2}
\end{equation}
where $\mathrm{vec}(\cdot)$ is the matrix vectorization operation,
and $\nu>0$ is the degrees of freedom hyperparameter of the imposed
priors. On this basis, we seek to devise an efficient means of inferring
the corresponding posterior distributions, given the available training
data. To this end, we postulate that the sought posteriors approximately
take the form of Student's-$t$ densities with means, \emph{diagonal}
covariance matrices, and degrees of freedom inferred from the data.
Hence, we have: 
\begin{equation}
q(\boldsymbol{\omega};\boldsymbol{\phi})=t(\boldsymbol{\omega}|\boldsymbol{\mu}_{\boldsymbol{\omega}},\mathrm{diag}(\boldsymbol{\sigma}_{\boldsymbol{\omega}}^{2}),\nu_{\boldsymbol{\omega}})\label{eq:post1}
\end{equation}

\begin{equation}
q(\boldsymbol{W};\boldsymbol{\phi})=t(\mathrm{vec}(\boldsymbol{W})|\boldsymbol{\mu}_{\boldsymbol{W}},\mathrm{diag}(\boldsymbol{\sigma}_{\boldsymbol{W}}^{2}),\nu_{\boldsymbol{W}})\label{eq:post2}
\end{equation}
where $\boldsymbol{\phi}=\{\boldsymbol{\mu}_{i},\boldsymbol{\sigma}_{i}^{2},\nu_{i}\}_{i\in\{\boldsymbol{\omega},\boldsymbol{W}\}}$,
and $\nu_{i}>0,\forall i$, for all model layers. This selection for
the form of the sought parameter posteriors allows for our model to
accommodate heavy-tailed underlying densities much better than the
usual Gaussian assumption, as we have also discussed in Section 2.2.

Then, to allow for inferring the sought posteriors in a computationally
efficient manner, we resort to variational Bayes \citep{attias}.
This consists in derivation of a family of variational posterior distributions
$q(.)$, which approximate the true posterior distributions that we
need to infer. In essence, this is effected by optimizing an appropriate
functional over the variational posterior, which measures model fit
to the training data. Hence, variational Bayes casts inference as
an approximate optimization problem, yielding a trade-off between
accuracy and computational scalability \citep{attias}. 

Specifically, to allow for reaping the most out of the heavy-tailed
assumptions of our model, we elect to perform variational Bayes by
minimizing the $t$-divergence between the sought approximate posterior
and the postulated joint density over the observed data and the model
latent variables \citep{tVB}. Thus, the proposed model training objective
becomes
\begin{equation}
\begin{aligned}q(\boldsymbol{\omega};\boldsymbol{\phi}),q(\boldsymbol{W};\boldsymbol{\phi}) & =\\
\underset{q(\cdot)}{=\mathrm{arg\,min}\;} & D_{t}\left(q(\boldsymbol{\omega};\boldsymbol{\phi}),q(\boldsymbol{W};\boldsymbol{\phi})||p(\boldsymbol{y};\boldsymbol{\omega},\boldsymbol{W})\right)
\end{aligned}
\label{eq:max}
\end{equation}

By application of  simple algebra, the expression of the $t$-divergence
in (\ref{eq:max}) yields
\begin{equation}
\begin{aligned}D_{t}\big(q(\boldsymbol{\omega};\boldsymbol{\phi}),q(\boldsymbol{W};\boldsymbol{\phi}) & ||p(\boldsymbol{y};\boldsymbol{\omega},\boldsymbol{W})\big)=\\
=D_{t}\left(q(\boldsymbol{\omega};\boldsymbol{\phi})||p(\boldsymbol{\omega})\right) & +D_{t}\left(q(\boldsymbol{W};\boldsymbol{\phi})||p(\boldsymbol{W})\right)\\
 & -\mathbb{E}_{\tilde{q}(\boldsymbol{\omega},\boldsymbol{W};\boldsymbol{\phi})}[\mathrm{log}p(\boldsymbol{y}|\boldsymbol{x})]
\end{aligned}
\label{eq:gen-obj}
\end{equation}
where $\tilde{q}(\boldsymbol{\omega},\boldsymbol{W};\boldsymbol{\phi})$
is the escort distribution of the sought posterior, and the $t$-divergence
terms pertaining to the parameters $\boldsymbol{\omega}$ and $\boldsymbol{W}$
are summed over all model layers. Then, following \citet{tVB}, and
based on (\ref{eq:prior1})-(\ref{eq:post2}), we obtain that the
$t$-divergence expressions in (\ref{eq:gen-obj}) can be written
in the following form:
\begin{equation}
\begin{aligned}D_{t}\left(q(\boldsymbol{\theta};\boldsymbol{\phi})||p(\boldsymbol{\theta})\right)=\sum_{i}\bigg\{ & \frac{\Psi_{qi}}{1-t}\left(1+\frac{1}{\nu_{\boldsymbol{\theta}}}\right)\\
-\frac{\Psi_{p}}{1-t} & \left(1+\frac{[\boldsymbol{\sigma}_{\boldsymbol{\theta}}^{2}]_{i}+[\boldsymbol{\mu_{\theta}}]_{i}^{2}}{\nu}\right)\bigg\}
\end{aligned}
\end{equation}
where $\boldsymbol{\theta}\in\{\boldsymbol{\omega},\mathrm{vec}(\boldsymbol{W})\}$,
$[\boldsymbol{\zeta}]_{i}$ is the $i$th element of a vector $\boldsymbol{\zeta}$,
we denote
\begin{equation}
\Psi_{qi}\triangleq\left(\frac{\mathrm{\Gamma}(\frac{\nu_{\boldsymbol{\theta}}+1}{2})}{\mathrm{\Gamma}(\frac{\nu_{\boldsymbol{\theta}}}{2})(\pi\nu_{\boldsymbol{\theta}})^{1/2}[\boldsymbol{\sigma}_{\boldsymbol{\theta}}]_{i}}\right)^{-\frac{2}{\nu_{\boldsymbol{\theta}}+1}}
\end{equation}
\begin{equation}
\Psi_{p}\triangleq\left(\frac{\mathrm{\Gamma}(\frac{\nu+1}{2})}{\mathrm{\Gamma}(\frac{\nu}{2})(\pi\nu)^{1/2}}\right)^{-\frac{2}{\nu+1}}
\end{equation}
 and the free hyperparameter $t$ is set as suggested in \citep{tVB},
yielding:
\begin{equation}
t=\frac{2}{1+\nu_{\boldsymbol{\theta}}}+1
\end{equation}

As we observe from the preceding discussion, the expectation of the
conditional log-likelihood of our model, $\mathbb{E}_{\tilde{q}(\boldsymbol{\omega},\boldsymbol{W};\boldsymbol{\phi})}[\mathrm{log}p(\boldsymbol{y}|\boldsymbol{x})]$,
is computed with respect to the escort distributions of the sought
posteriors. Therefore, \emph{at training time,} the banks of the employed
feature functions (i.e., the samples of their parameters, $\{\boldsymbol{\omega}_{s}\}_{s=1}^{S}$),
must be drawn \emph{from the escort distributions of the derived posteriors. }

Turning to the employed mixing weight matrices, $\boldsymbol{W}$,
our consideration of them being latent variables with an inferred
posterior perplexes computation of the expressions (\ref{eq:def-1})-(\ref{eq:def-3});
indeed, it requires that we compute appropriate posterior expectations
of these expressions w.r.t. $\boldsymbol{W}$. To circumvent this
problem, we draw multiple samples of the mixing weight matrices, $\{\boldsymbol{W}_{s}\}_{s=1}^{S}$,
\emph{in an MC fashion,} and \emph{average over the corresponding
outcomes to compute the model output.} \emph{At training time, }Eq.
(\ref{eq:gen-obj}) implies that these samples must also be drawn
\emph{from the escort distributions of the derived posteriors.} 

Based on the previous results, and following \citep{tVB}, we can
easily obtain the expressions of these escort distributions of the
derived posteriors, that we need to sample from at training time.
Specifically, it is easy to show that these escort distributions yield
a factorized form, that reads:
\begin{equation}
\begin{aligned}\tilde{q}(\boldsymbol{\theta};\boldsymbol{\phi})=t\left(\boldsymbol{\theta}|\boldsymbol{\mu_{\theta}},\frac{\nu_{\boldsymbol{\theta}}}{\nu_{\boldsymbol{\theta}}+2}\mathrm{diag}(\boldsymbol{\sigma}_{\boldsymbol{\theta}}^{2}),\nu_{\boldsymbol{\theta}}+2\right)\\
\forall\boldsymbol{\theta}\in\{\boldsymbol{\omega},\mathrm{vec}(\boldsymbol{W})\}
\end{aligned}
\label{eq:esc-post}
\end{equation}

Notably, our inference algorithm yields an MC estimator of the proposed
D$t$BKS model. Unfortunately, MC estimators are notorious for their
vulnerability to unacceptably high variance. In this work, we resolve
these issues by adopting the reparameterization trick of \citep{aevb},
adapted to the $t$-exponential family. This trick consists in a smart
reparameterization of the MC samples, $\{\boldsymbol{\theta}_{s}\}_{s=1}^{S}$,
drawn from a distribution $\boldsymbol{\theta}\sim q(\boldsymbol{\theta};\boldsymbol{\phi})$;
this is obtained via a differentiable transformation $\boldsymbol{g}_{\boldsymbol{\phi}}(\boldsymbol{\epsilon})$
of an (auxiliary) random variable $\boldsymbol{\epsilon}$ with low
variance: 
\begin{equation}
\boldsymbol{\theta}=\boldsymbol{g}_{\boldsymbol{\phi}}(\boldsymbol{\epsilon})\;\;\;\mathrm{with}\;\;\;\boldsymbol{\epsilon}\sim p(\boldsymbol{\epsilon})
\end{equation}

In our case, the smart reparameterization of the MC samples drawn
from the Student's-$t$ escort densities (\ref{eq:esc-post}) yields
the expression: 
\begin{equation}
\boldsymbol{\theta}_{s}=\boldsymbol{\theta}(\boldsymbol{\epsilon}_{s})=\boldsymbol{\mu}_{\boldsymbol{\theta}}+\left(\frac{\nu_{\boldsymbol{\theta}}}{\nu_{\boldsymbol{\theta}}+2}\right)^{1/2}\boldsymbol{\sigma}_{\boldsymbol{\theta}}\boldsymbol{\epsilon}_{s}\label{eq:trick}
\end{equation}
where $\boldsymbol{\epsilon}_{s}$ is random Student's-$t$ noise
with unitary variance:
\begin{equation}
\boldsymbol{\epsilon}_{s}\sim t(\boldsymbol{0},\boldsymbol{I},\nu_{\boldsymbol{\theta}}+2)
\end{equation}

On this basis,\emph{ at training time}, we replace the samples of
both the weight matrices, $\boldsymbol{W}$, as well as the feature
function parameters, $\boldsymbol{\omega}$, with the expression (\ref{eq:trick}),
where sampling is performed w.r.t. the low-variance random variable
$\boldsymbol{\epsilon}$. Then, the resulting (reparameterized) $t$-divergence
objective (\ref{eq:gen-obj}) can be minimized by means of any off-the-shelf
stochastic optimization algorithm, yielding low variance estimators.
To this end, in this work we utilize AdaM \citep{adam}. We initialize
the sought posterior hyperparameters by setting them equal to the
hyperparameters of the imposed priors.

\subsection{Inference Algorithm}

Having obtained a training algorithm for our proposed D$t$BKS model,
we can now proceed to elaborate on how inference is performed using
our method. To this end, it is needed that we compute the \emph{posterior
expectation} of the model's output, as usual when performing Bayesian
inference. Thus, \emph{at inference time}, we need to \emph{draw samples
from the derived posteriors}, (\ref{eq:post1}) and (\ref{eq:post2}),
in an MC fashion. This entails drawing from the posteriors, at each
layer, of a set comprising $S$ samples of: (i) the vectors $\boldsymbol{\omega}$
that parameterize the employed feature functions; and (ii) the mixing
weight matrices, $\boldsymbol{W}$, used to combine the outputs of
the drawn banks of feature functions. Note that this is in contrast
to the training algorithm of D$t$BKS, where the use of the $t$-divergence
objective (\ref{eq:gen-obj}) gives rise to the requirement of drawing
from the associated \emph{escort distributions} (\ref{eq:esc-post}),
while the need of training reliable estimators requires utilization
of the reparameterization trick.

\section{Experimental Evaluation}

In this Section, we perform a thorough experimental evaluation of
our proposed D$t$BKS model. We provide a quantitative assessment
of the efficacy, the effectiveness, and the computational efficiency
of our approach, combined with deep qualitative insights into few
of its key performance characteristics. To this end, we consider several
benchmarks from the UCI machine learning repository (UCI-Rep) \citep{uci}
that pertain to both regression and classification tasks, as well
as the well-known InfiMNIST classification benchmark \citep{mnist8m}.
The considered datasets, as well as their main characteristics (i.e.,
their number of training examples, $N$, and input dimensionality,
$\delta$) are summarized in Table 1.

\begin{table*}
\caption{Characteristics of the considered datasets.}

\centering{}%
\begin{tabular}{|c|c|c|c|c|}
\hline 
Dataset & $N$ & $\delta$ & Task & Performance Metric\tabularnewline
\hline 
\hline 
Boston Housing & 506 & 13 & Regression & RMSE\tabularnewline
\hline 
Concrete & 1030 & 8 & Regression & RMSE\tabularnewline
\hline 
Energy & 768 & 8 & Regression & RMSE\tabularnewline
\hline 
Power Plant & 9568 & 4 & Regression & RMSE\tabularnewline
\hline 
Protein & 45730 & 9 & Regression & RMSE\tabularnewline
\hline 
Wine (White) & 4898 & 11 & Regression & RMSE\tabularnewline
\hline 
Wine (Red) & 1588 & 11 & Regression & RMSE\tabularnewline
\hline 
\hline 
Breast Cancer Diagnostic (wdbc) & 569 & 30 & Classification & Misclassification Rate\tabularnewline
\hline 
ISOLET & 7797 & 617 & Classification & Misclassification Rate\tabularnewline
\hline 
Gas Sensor & 13910 & 128 & Classification & Misclassification Rate\tabularnewline
\hline 
Parkinson's & \multirow{2}{*}{197} & \multirow{2}{*}{22} & \multirow{2}{*}{Classification} & \multirow{2}{*}{Misclassification Rate}\tabularnewline
(Oxford Parkinson's Disease Detection Dataset) &  &  &  & \tabularnewline
\hline 
Spam & 4601 & 56 & Classification & Misclassification Rate\tabularnewline
\hline 
LSVT Voice Rehabilitation & 126 & 310 & Classification & Misclassification Rate\tabularnewline
\hline 
InfiMNIST & 8+ Million & 784 & Classification & Misclassification Rate\tabularnewline
\hline 
\end{tabular}
\end{table*}

With the exception of the ISOLET dataset from UCI-Rep, as well as
InfiMNIST, the rest of the considered benchmarks do not provide a
split into training and test sets. In these cases, we account for
this lack by running our experiments 20 times, with different random
data splits into training and test sets, and compute performance means
and standard deviations; we use a randomly selected 90\% of the data
for model training, and the rest for evaluation purposes. In the case
of regression tasks, we use the root mean square error (RMSE) as our
performance metric; we employ the misclassification rate for the considered
classification benchmarks.

To obtain some comparative results, we also evaluate an existing alternative
approach for Bayesian inference of deep network nonlinearities, namely
deep Gaussian processes (DGPs) \citep{vb-weights}. Indeed, the DGP
is the existing type of Bayesian deep learning approaches that is
most closely related to our approach; this is the case, since DGP
does also allow for inferring the employed nonlinearities, but under
a completely different rationale. Specifically, we evaluate the most
recent variant of DGPs, presented in \citep{dgp}, which allows for
the model to efficiently scale to large datasets. Finally, we also
provide comparisons to a state-of-the-art Deep Learning approach,
namely a Dropout network \citep{dropout}, as well as a baseline approach,
namely SVMs using both linear and RBF kernels.

In all cases, our specification of the priors imposed on D$t$BKS
considers a low value for the degrees of freedom hyperparameter, $\nu=2.1$.
In the case of regression tasks, we employ a noise variance equal
to $\sigma_{y}^{2}=\mathrm{exp}(-2)$. Turning to the selection of
the form of the drawn feature functions, $\xi$, we consider a simple
\emph{trigonometric }formulation, which is inspired from the theory
of random Fourier projections of RBF kernels \citep{bochner}. Specifically,
we postulate $\xi(\boldsymbol{x};\boldsymbol{\omega})=\frac{1}{2}\mathrm{cos}(\boldsymbol{\omega}^{T}\boldsymbol{x})+\frac{1}{2}\mathrm{sin}(\boldsymbol{\omega}^{T}\boldsymbol{x})$.
Note that under this selection for the form of $\xi(\boldsymbol{x};\boldsymbol{\omega})$,
and setting $\nu_{\boldsymbol{\omega}}\rightarrow\infty$ and $\nu_{\boldsymbol{W}}\rightarrow\infty$,
D$t$BKS reduces to the DGP variant of \citep{dgp} with a postulated
RBF kernel. For computational efficiency, we limit the number of drawn
samples to 100, during both D$t$BKS training and inference on the
test data. AdaM is run with the default hyperparameter values.

DGP is evaluated considering multiple selections of the number of
Gaussian processes per layer, as well as the number of layers, using
RBF kernels and arc-cosine kernels \citep{dgp}; in each experimental
case, we report results pertaining to the best-performing DGP configuration.
Similar is the case for Dropout networks, which are evaluated considering
multiple alternatives for the number of layers and the output size
of each hidden layer; we employ ReLU nonlinearities \citep{relu}.

Our deep learning source code has been developed in Python, using
the Tensorflow library \citep{tensorflow2015-whitepaper}; it can
be found on \url{https://github.com/Partaourides/DtBKS}.\textbf{
}We have also made use of a DGP implementation provided by M. Filippone\footnote{\url{https://github.com/mauriziofilippone/deep_gp_random_features}}.
The evaluation of SVM-type algorithms was performed by utilizing Python's
scikit-learn toolbox \citep{scikit-learn}. We run our experiments
on an Intel Xeon 2.5GHz server with 64GB RAM and an NVIDIA Tesla K40
GPU.

\begin{table*}
\caption{Obtained performance for best model configuration (RMSE for regression
tasks, misclassification rate for classification tasks; the lower
the better).}

\centering{}%
\begin{tabular}{|c||c|c|c|c|c|}
\hline 
Dataset & D$t$BKS & DGP & Dropout & Linear SVM & RBF-kernel SVM\tabularnewline
\hline 
\hline 
\multirow{2}{*}{Boston Housing} & $0.2939\pm0.04$ & $0.3897\pm0.1$ & $0.2516\pm0.06$ & $0.5027\pm0.1$ & $0.3471\pm0.1$\tabularnewline
 & ($L=2$,$\eta=3$) & ($L=2$,$\eta=3$) & ($L=2$,$\eta=123$)  &  & \tabularnewline
\hline 
\multirow{2}{*}{Concrete} & $0.3213\pm0.02$ & $0.4501\pm0.03$ & $0.3228\pm0.03$ & $0.64\pm0.04$ & $0.3957\pm0.03$\tabularnewline
 &  ($L=2$,$\eta=5$) & ($L=2$,$\eta=5$) & ($L=2$,$\eta=163$)  &  & \tabularnewline
\hline 
\multirow{2}{*}{Energy} & $0.1285\pm0.01$ & $0.1636\pm0.02$ & $0.1322\pm0.01$ & $0.3183\pm0.03$ & $0.25\pm0.03$\tabularnewline
 &  ($L=2$,$\eta=6$) & ($L=2$,$\eta=6$) & ($L=2$,$\eta=175$)  &  & \tabularnewline
\hline 
\multirow{2}{*}{Power Plant} & $0.2366\pm0.01$ & $0.2401\pm0.01$  & $0.2236\pm0.01$ & $0.2671\pm0.01$ & $0.2399\pm0.01$\tabularnewline
 & ($L=3$,$\eta=4$) & ($L=3$,$\eta=4$) & ($L=3$,$\eta=200$)  &  & \tabularnewline
\hline 
\multirow{2}{*}{Protein} & $0.6113\pm0.01$ & $0.6734\pm0.01$ & $0.7453\pm0.01$ & $0.8673\pm0.01$ & $0.7648\pm0.01$\tabularnewline
 & ($L=2$,$\eta=9$) & ($L=2$,$\eta=9$) & ($L=2$,$\eta=200$)  &  & \tabularnewline
\hline 
\multirow{2}{*}{Wine (White)} & $0.7684\pm0.02$ & $0.8072\pm0.02$ & $0.7609\pm0.02$ & $0.8551\pm0.02$ & $0.7751\pm0.02$\tabularnewline
 & ($L=2$,$\eta=11$) & ($L=2$,$\eta=11$) & ($L=2$,$\eta=200$)  &  & \tabularnewline
\hline 
\multirow{2}{*}{Wine (Red)} & $0.7564\pm0.04$ & $0.7791\pm0.04$ & $0.7570\pm0.05$ & $0.8064\pm0.05$ & $0.7693\pm0.05$\tabularnewline
 & ($L=2$,$\eta=5$) & ($L=2$,$\eta=5$) & ($L=2$,$\eta=145$)  &  & \tabularnewline
\hline 
\multirow{2}{*}{Breast Cancer Diagnostic (wdbc)} & $0.0116\pm0.01$ & $0.0116\pm0.01$ & $0.0710\pm0.06$ & $0.0304\pm0.02$ & $0.025\pm0.02$\tabularnewline
 & ($L=2$,$\eta=21$) & ($L=2$,$\eta=21$) & ($L=2$,$\eta=170$)  &  & \tabularnewline
\hline 
\multirow{2}{*}{ISOLET} & $0.055\pm NA$ & $0.0654\pm NA$ & $0.1256\pm NA$ & $0.055\pm NA$ & $0.063\pm NA$\tabularnewline
 & ($L=2$,$\eta=205$) & ($L=2$,$\eta=205$) & ($L=2$,$\eta=133$)  &  & \tabularnewline
\hline 
\multirow{2}{*}{Gas Sensor} & $0.0136\pm0.002$ & $0.0094\pm0.002$ & $0.0688\pm0.07$ & $0.0159\pm0.001$ & $0.0168\pm0.003$\tabularnewline
 & ($L=3$,$\eta=106$) & ($L=3$,$\eta=106$) & ($L=3$,$\eta=183$)  &  & \tabularnewline
\hline 
\multirow{2}{*}{Parkinson's} & $0.0658\pm0.05$ & $0.0842\pm0.05$ & $0.0976\pm0.09$ & $0.1342\pm0.07$ & $0.1079\pm0.04$\tabularnewline
 & ($L=2$,$\eta=15$) & ($L=2$,$\eta=15$) & ($L=2$,$\eta=168$)  &  & \tabularnewline
\hline 
\multirow{2}{*}{Spam} & $0.0543\pm0.01$ & $0.0517\pm0.01$ & $0.1629\pm0.03$ & $0.0777\pm0.01$ & $0.0666\pm0.01$\tabularnewline
 & ($L=2$,$\eta=46$) & ($L=2$,$\eta=46$) & ($L=2$,$\eta=182$)  &  & \tabularnewline
\hline 
\multirow{2}{*}{LSVT Voice Rehabilitation} & $0.1375\pm0.07$ & $0.3250\pm0.11$ & $0.3782\pm0.17$ & $0.2583\pm0.09$ & $0.1667\pm0.08$\tabularnewline
 & ($L=2$,$\eta=51$) & ($L=2$,$\eta=51$) & ($L=2$,$\eta=116$)  &  & \tabularnewline
\hline 
\multirow{2}{*}{InfiMNIST} & $0.0093\pm NA$ & $0.0096\pm NA$ & $0.0096\pm NA$ & $0.25\pm NA$ & $0.25\pm N$A\tabularnewline
 & ($L=4$,$\eta=100$) & ($L=4$,$\eta=100$) & ($L=4$,$\eta=113$)  &  & \tabularnewline
\hline 
\end{tabular}
\end{table*}

\begin{figure*}
\begin{centering}
\subfloat[]{\begin{centering}
\includegraphics[scale=0.51]{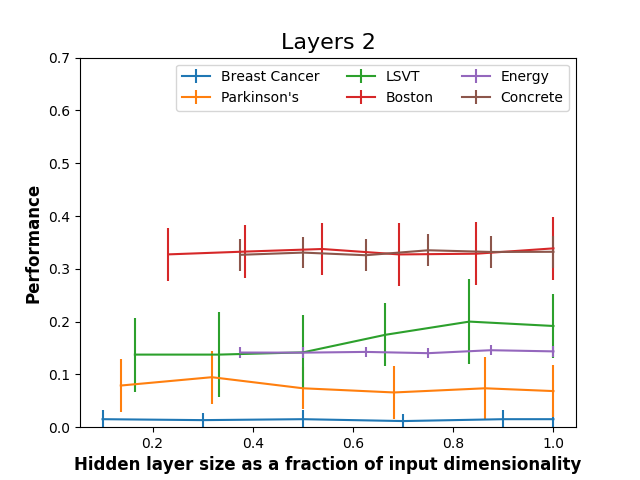}
\par\end{centering}
}\hfill{}\subfloat[]{\begin{centering}
\includegraphics[scale=0.51]{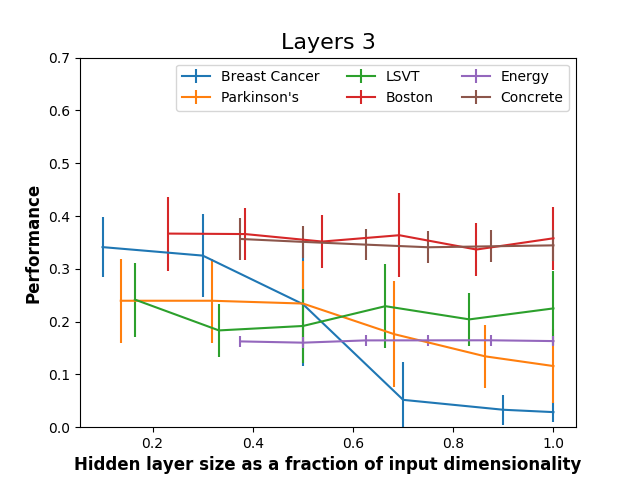}
\par\end{centering}
}\hfill{}\subfloat[]{\begin{centering}
\includegraphics[scale=0.51]{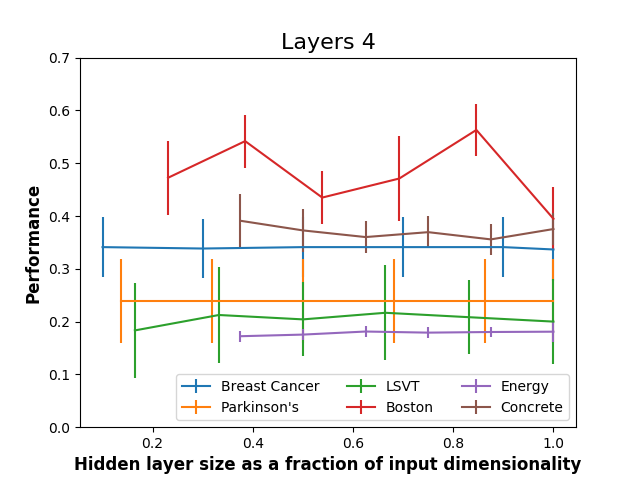}
\par\end{centering}
}
\par\end{centering}
\centering{}\caption{D$t$BKS performance fluctuation with the number of layers, $L$,
and the output size of each hidden layer, $\eta$ (as a fraction of
input dimensionality, $\delta$): (a) $L=2$; (b) $L=3$; (c) $L=4$.
Performance metrics are the RMSE for regression tasks, and the misclassification
rate for classification tasks (the lower the better).}
\end{figure*}

\subsection{Comparative Results}

We begin our exposition by providing the best empirical performance
of our method, and showing how it compares to the alternatives. These
outcomes have been obtained by experimenting with different selections
for the number of layers, $L$, and the output size of each hidden
layer, $\eta$ (i.e., for $l\in\{1,\dots,L-1\}$). Our results are
outlined in Table 2; in all cases, we provide therein (in parentheses)
the model configurations that obtained the reported (best empirical)
performance\footnote{This selection was performed by means of leave-one-out cross-validation,
considering $L\in\{2,3,4,5\}$ and $\eta\in\{\lceil\delta/4\rceil,\lceil\delta/2\rceil,\lceil3\delta/4\rceil,\delta\}$.}. 

We observe that our approach outperforms DGP in the considered regression
benchmarks; in all cases, these empirical performance differences
are found to be statistically significant, by running the paired Student's-$t$
test. On the other hand, D$t$BKS outperforms DGP in only three out
of the seven considered classification benchmarks, with statistically
significant differences (according to the paired Student's-$t$ test),
while yielding comparable outcomes in the rest. In addition, D$t$BKS
outperforms Dropout in all the considered classification benchmarks;
the paired Student's-$t$ test shows that these empirical performance
differences are statistically significant. The only exception to this
finding is InfiMNIST, where the results are essentially comparable.
On the other hand, D$t$BKS significantly outperforms Dropout in the
Protein regression benchmark, while yielding comparable performance
in the rest considered regression tasks (according to the outcomes
of the paired Student's-$t$ test). Finally, both baseline SVM model
configurations are completely outperformed by D$t$BKS, in all cases.

\subsection{Further Investigation}

Further, it is interesting to provide a feeling of how D$t$BKS model
performance changes with the selection of the number of layers, $L$,
and the dimensionality of each hidden layer, $\eta$ (i.e., for $l\in\{1,\dots,L-1\}$).
To examine these aspects, in Figs. 3(a)-(c) we plot model performance
fluctuation with $\eta$, setting the number of layers equal to $L=2,3$,
and $4$, respectively, for few characteristic experimental cases
comprising limited training data. As we observe, D$t$BKS performance
is significantly affected by both these selections. Note also that
the associated performance fluctuation patterns of D$t$BKS are quite
different among the illustrated examples. These findings are congruent
with the behavior of all existing state-of-the-art deep learning approaches.
It is also important to mention the high standard deviation of the
observed performances in some cases where we set $L=4$; we attribute
this unstable behavior to overfitting due to insufficient training
data.

\subsection{Are $t$-Exponential Bayesian Kitchen Sinks More Potent Than Random
Kitchen Sinks?}

Finally, it is extremely interesting to examine how beneficial it
is for D$t$BKS to infer a posterior distribution over the (random
variables that parameterize the) employed feature functions, instead
of using a simple, manually selected density. To examine this aspect,
we repeat our experiments by drawing the vectors $\boldsymbol{\omega}$,
that parameterize the feature functions, $\xi$, from the postulated
simple priors, $p(\boldsymbol{\omega})$. Hence, we adopt an RKS-type
rationale in drawing the feature functions, $\xi$, as opposed to
utilizing the inferred posteriors, $q(\boldsymbol{\omega})$ {[}or
their corresponding escort distributions, $\tilde{q}(\boldsymbol{\omega})$,
during training{]}. 

Our findings are provided in Table 3; these results correspond to
selections of the number of layers, $L$, and the output size, $\eta$,
similar to the values reported in Table 2. Our empirical evidence
is quite conspicuous: (i) merely drawing the postulated nonlinearities
from a simple prior, yet inferring a Student's-$t$ posterior over
the mixing weights, $\boldsymbol{W}$, as discussed previously, yields
notably competitive performance; (ii) inferring posteriors over the
nonlinearities, under the discussed D$t$BKS rationale, gives a statistically
significant boost to the obtained modeling performance, except for
\emph{Power Plant }and \emph{LSVT, }where we reckon that overfitting
is induced (due to insufficient training data availability).

\begin{table*}
\caption{D$t$BKS performance when replacing $t$-Exponential Bayesian Kitchen
Sinks with Random Kitchen Sinks. Performance metrics are the RMSE
for regression tasks, and the misclassification rate for classification
tasks (the lower the better).}

\centering{}%
\begin{tabular}{|c||c||c|}
\hline 
Dataset & Performance & Comparison to full D$t$BKS model (from Table 2)\tabularnewline
\hline 
\hline 
Boston Housing & $0.3199\pm0.04$ & $0.2939\pm0.04$\tabularnewline
\hline 
Concrete & $0.3586\pm0.04$ & $0.3213\pm0.02$\tabularnewline
\hline 
Energy & $0.1494\pm0.01$ & $0.1285\pm0.01$\tabularnewline
\hline 
Power Plant & $0.2301\pm0.01$ & $0.2366\pm0.01$\tabularnewline
\hline 
Protein & $0.7110\pm0.01$  & $0.6113\pm0.01$\tabularnewline
\hline 
Wine (White) & $0.7787\pm0.02$ & $0.7684\pm0.02$\tabularnewline
\hline 
Wine (Red) & $0.7720\pm0.04$ & $0.7564\pm0.04$\tabularnewline
\hline 
\hline 
Breast Cancer Diagnostic (wdbc) & $0.0188\pm0.02$ & $0.0116\pm0.01$\tabularnewline
\hline 
ISOLET & $0.2245\pm NA$ & $0.0552\pm NA$\tabularnewline
\hline 
Gas Sensor & $0.0175\pm0.003$ & $0.0136\pm0.002$\tabularnewline
\hline 
Parkinson's & $0.0895\pm0.06$ & $0.0658\pm0.05$\tabularnewline
\hline 
Spam & $0.0748\pm0.01$ & $0.0543\pm0.01$\tabularnewline
\hline 
LSVT Voice Rehabilitation & $0.1208\pm0.04$ & $0.1375\pm0.07$\tabularnewline
\hline 
InfiMNIST & $0.0603\pm NA$ & $0.0093\pm NA$\tabularnewline
\hline 
\end{tabular}
\end{table*}

\subsection{Computational Complexity}

Another significant aspect that affects the efficacy of a machine
learning technique is its computational complexity. To investigate
this aspect, we scrutinize the derived D$t$BKS algorithm, both regarding
its asymptotic behavior, as well as in terms of its total computational
costs. Our observations can be summarized as follows: For the model
configurations yielding the performance statistics of Table 1, D$t$BKS
takes on average 4 times longer than Dropout \emph{per algorithm iteration},
probably due to the entailed $\Gamma(\cdot)$ functions in (16), and
their derivatives; DGP takes on average 2 times longer than Dropout.
On the other hand, D$t$BKS training converges much faster than all
the considered competitors. These differences are so immense that,
as an outcome, the \emph{total time} required \emph{by all the evaluated
methods} is of the same order of magnitude. Indeed, we typically observe
that D$t$BKS takes much less time than DGP, and usually not much
longer than Dropout; these outcomes are summarized in Table 4. Hence,
we deduce that D$t$BKS yields the observed predictive performance
improvement without undermining computational efficiency and scalability. 

\begin{table*}
\caption{Wall-clock times of the evaluated deep learning approaches (in minutes).}

\centering{}%
\begin{tabular}{|c||c|c|c|c|}
\hline 
Dataset & D$t$BKS & RKS & DGP & Dropout\tabularnewline
\hline 
\hline 
\multirow{1}{*}{Boston Housing} & 8 & 4 & 11 & 12\tabularnewline
\hline 
\multirow{1}{*}{Concrete} & 20 & 7 & 48 & 14\tabularnewline
\hline 
\multirow{1}{*}{Energy} & 23 & 8 & 166 & 22\tabularnewline
\hline 
\multirow{1}{*}{Power Plant} & 94 & 43 & 91 & 49\tabularnewline
\hline 
\multirow{1}{*}{Protein} & 74 & 48 & 35 & 43\tabularnewline
\hline 
\multirow{1}{*}{Wine (White)} & 20 & 14 & 13 & 10\tabularnewline
\hline 
\multirow{1}{*}{Wine (Red)} & 8 & 7 & 10 & 6\tabularnewline
\hline 
\multirow{1}{*}{Breast Cancer Diagnostic (wdbc)} & 14 & 5 & 4 & 15\tabularnewline
\hline 
\multirow{1}{*}{ISOLET} & 195 & 31 & 109 & 20\tabularnewline
\hline 
\multirow{1}{*}{Gas Sensor} & 97 & 35 & 39 & 48\tabularnewline
\hline 
\multirow{1}{*}{Parkinson's} & 9 & 6 & 7 & 20\tabularnewline
\hline 
\multirow{1}{*}{Spam} & 42 & 22 & 23 & 13\tabularnewline
\hline 
\multirow{1}{*}{LSVT Voice Rehabilitation} & 11 & 5 & 1 & 3\tabularnewline
\hline 
\multirow{1}{*}{InfiMNIST} & 650 & 1165 & 489 & 238\tabularnewline
\hline 
\end{tabular}
\end{table*}

\section{Conclusions}

In this paper, we introduced a fresh view towards deep learning, which
consists in postulating banks of randomly drawn nonlinearities at
each model layer. To alleviate the burden of having to manually specify
the distribution these nonlinear feature functions are drawn from,
we elected to infer them in a Bayesian sense. This also renders our
model more robust to scenarios dealing with limited or sparse training
data availability.

In this context, we postulated that the sought posteriors constitute
multivariate Student's-$t$ densities. This assumption allows for
our model to better cope with heavy-tailed underlying densities; these
are quite common in real-world data modeling scenarios, yet they cannot
be captured sufficiently enough by the usual Gaussian assumptions.
Then, to allow for reaping the most out of the heavy tails of Student's-$t$
densities, we performed approximate Bayesian inference for our model
under a novel objective function construction. This was based on a
$t$-divergence functional, which better accommodates heavy-tailed
densities.

We exhaustively evaluated our approach using challenging benchmark
datasets; we offered thorough insights into its key performance characteristics.
This way, we illustrated that our proposed approach outperforms the
existing alternatives in terms of predictive accuracy, without undermining
the overall computational scalability, both in terms of training time
and of prediction generation time. We also showed that data-driven
inference of a posterior distribution from which we can draw the employed
banks of nonlinearities yields better results than drawing from a
simple prior.

One direction for further research concerns postulating nonelliptical
latent variable densities, which can account for skewness in a fashion
similar, e.g., to \citep{asydgm,pami10}. Introduction of a solid
means of capturing conditional heteroscedasticity in modeled sequential
data, in a fashion similar, e.g., to \citep{gpmch}, is also a challenge
of immense interest. On a different vein, we must emphasize that our
approach is not capable of modeling spatial dynamics and dependencies
the way, e.g., convolutional networks do. This is similar to related
approaches, such as the DGP model and Dropout networks, which are
also not designed with such tasks in mind. However, enabling such
capabilities in the context of our D$t$BKS framework would be extremely
auspicious for the model performance in the context of real-world
applications, dealing with challenging image data. Hence, addressing
these challenges and examining the associated opportunities remains
to be explored in our future work.

\bibliographystyle{model2-names}
\bibliography{tbk}

\begin{thebibliography}{44}
\expandafter\ifx\csname natexlab\endcsname\relax\def\natexlab#1{#1}\fi
\providecommand{\url}[1]{\texttt{#1}}
\providecommand{\href}[2]{#2}
\providecommand{\path}[1]{#1}
\providecommand{\DOIprefix}{doi:}
\providecommand{\ArXivprefix}{arXiv:}
\providecommand{\URLprefix}{URL: }
\providecommand{\Pubmedprefix}{pmid:}
\providecommand{\doi}[1]{\href{http://dx.doi.org/#1}{\path{#1}}}
\providecommand{\Pubmed}[1]{\href{pmid:#1}{\path{#1}}}
\providecommand{\bibinfo}[2]{#2}
\ifx\xfnm\relax \def\xfnm[#1]{\unskip,\space#1}\fi
\bibitem[{Abadi et~al.(2015)}]{tensorflow2015-whitepaper}
\bibinfo{author}{Abadi, M.}, et~al., \bibinfo{year}{2015}.
\newblock \bibinfo{title}{{TensorFlow}: Large-scale machine learning on
  heterogeneous systems}.
\newblock \URLprefix \url{http://tensorflow.org/}. \bibinfo{note}{software
  available from tensorflow.org}.
\bibitem[{Archambeau and Verleysen(2007)}]{vbsmm}
\bibinfo{author}{Archambeau, C.}, \bibinfo{author}{Verleysen, M.},
  \bibinfo{year}{2007}.
\newblock \bibinfo{title}{Robust {Bayesian} clustering}.
\newblock \bibinfo{journal}{Neural Networks} \bibinfo{volume}{20},
  \bibinfo{pages}{129--138}.
\bibitem[{Asuncion and Newman(2007)}]{uci}
\bibinfo{author}{Asuncion, A.}, \bibinfo{author}{Newman, D.},
  \bibinfo{year}{2007}.
\newblock \bibinfo{title}{{UCI} machine learning repository}.
\newblock \URLprefix
  \url{http://www.ics.uci.edu/$\sim$mlearn/{MLR}epository.html}.
\bibitem[{Attias(2000)}]{attias}
\bibinfo{author}{Attias, H.}, \bibinfo{year}{2000}.
\newblock \bibinfo{title}{A variational {Bayesian} framework for graphical
  models}, in: \bibinfo{booktitle}{Proc. NIPS'00}.
\bibitem[{Bui et~al.(2016)Bui, Hern\'andez-Lobato, Hern\'andez-Lobato, Li and
  Turner}]{ep}
\bibinfo{author}{Bui, T.}, \bibinfo{author}{Hern\'andez-Lobato, D.},
  \bibinfo{author}{Hern\'andez-Lobato, J.}, \bibinfo{author}{Li, Y.},
  \bibinfo{author}{Turner, R.}, \bibinfo{year}{2016}.
\newblock \bibinfo{title}{{Deep Gaussian Processes for Regression using
  Approximate Expectation Propagation}}.
\newblock \bibinfo{journal}{Proc. ICML} .
\bibitem[{Bui et~al.(2015)Bui, Hern\'andez-Lobato, Li, Hern\'andez-Lobato and
  Turner}]{probbp}
\bibinfo{author}{Bui, T.D.}, \bibinfo{author}{Hern\'andez-Lobato, J.M.},
  \bibinfo{author}{Li, Y.}, \bibinfo{author}{Hern\'andez-Lobato, D.},
  \bibinfo{author}{Turner, R.E.}, \bibinfo{year}{2015}.
\newblock \bibinfo{title}{{Training Deep Gaussian Processes using Stochastic
  Expectation Propagation and Probabilistic Backpropagation}}, in:
  \bibinfo{booktitle}{Proc. NIPS}.
\bibitem[{Chatzis(2010)}]{pami10}
\bibinfo{author}{Chatzis, S.}, \bibinfo{year}{2010}.
\newblock \bibinfo{title}{{Hidden Markov Models with Nonelliptically Contoured
  State Densities}}.
\newblock \bibinfo{journal}{IEEE Trans. Pattern Analysis and Machine
  Intelligence} \bibinfo{volume}{32}, \bibinfo{pages}{2297--2304}.
\bibitem[{Chatzis et~al.(2009)Chatzis, Kosmopoulos and Varvarigou}]{shmm}
\bibinfo{author}{Chatzis, S.}, \bibinfo{author}{Kosmopoulos, D.},
  \bibinfo{author}{Varvarigou, T.}, \bibinfo{year}{2009}.
\newblock \bibinfo{title}{Robust sequential data modeling using an outlier
  tolerant hidden {Markov} model}.
\newblock \bibinfo{journal}{IEEE Trans. Pattern Analysis and Machine
  Intelligence} \bibinfo{volume}{31}, \bibinfo{pages}{1657--1669}.
\bibitem[{Chatzis and Kosmopoulos(2011)}]{vbshmm}
\bibinfo{author}{Chatzis, S.P.}, \bibinfo{author}{Kosmopoulos, D.},
  \bibinfo{year}{2011}.
\newblock \bibinfo{title}{{A Variational Bayesian Methodology for Hidden Markov
  Models utilizing Student's-t Mixtures}}.
\newblock \bibinfo{journal}{Pattern Recognition} \bibinfo{volume}{44},
  \bibinfo{pages}{295--306}.
\bibitem[{Chatzis and Kosmopoulos(2015)}]{tnnls15}
\bibinfo{author}{Chatzis, S.P.}, \bibinfo{author}{Kosmopoulos, D.},
  \bibinfo{year}{2015}.
\newblock \bibinfo{title}{{A Latent Manifold Markovian Dynamics Gaussian
  Process}}.
\newblock \bibinfo{journal}{IEEE Transactions on Neural Networks and Learning
  Systems} \bibinfo{volume}{25}, \bibinfo{pages}{70--83}.
\bibitem[{Cutajar et~al.(2017)Cutajar, Bonilla, Michiardi and Filippone}]{dgp}
\bibinfo{author}{Cutajar, K.}, \bibinfo{author}{Bonilla, E.V.},
  \bibinfo{author}{Michiardi, P.}, \bibinfo{author}{Filippone, M.},
  \bibinfo{year}{2017}.
\newblock \bibinfo{title}{Random feature expansions for deep {Gaussian}
  processes}, in: \bibinfo{booktitle}{Proc. ICML}.
\bibitem[{Damianou and Lawrence(2013)}]{vb-weights}
\bibinfo{author}{Damianou, A.}, \bibinfo{author}{Lawrence, N.},
  \bibinfo{year}{2013}.
\newblock \bibinfo{title}{{Deep Gaussian Processes}}, in:
  \bibinfo{booktitle}{Proc. AISTATS}.
\bibitem[{Ding et~al.(2011)Ding, Vishwanathan and Qi}]{tVB}
\bibinfo{author}{Ding, N.}, \bibinfo{author}{Vishwanathan, S.N.},
  \bibinfo{author}{Qi, Y.}, \bibinfo{year}{2011}.
\newblock \bibinfo{title}{$t$-divergence based approximate inference}, in:
  \bibinfo{booktitle}{Proc. NIPS}.
\bibitem[{He et~al.(2016)He, Zhang, Ren and Sun}]{he}
\bibinfo{author}{He, K.}, \bibinfo{author}{Zhang, X.}, \bibinfo{author}{Ren,
  S.}, \bibinfo{author}{Sun, J.}, \bibinfo{year}{2016}.
\newblock \bibinfo{title}{Deep residual learning for image recognition}, in:
  \bibinfo{booktitle}{Proc. CVPR}.
\bibitem[{Kingma and Ba(2015)}]{adam}
\bibinfo{author}{Kingma, D.}, \bibinfo{author}{Ba, J.}, \bibinfo{year}{2015}.
\newblock \bibinfo{title}{Adam: A method for stochastic optimization}, in:
  \bibinfo{booktitle}{Proc. ICLR}.
\bibitem[{Kingma and Welling(2014)}]{aevb}
\bibinfo{author}{Kingma, D.}, \bibinfo{author}{Welling, M.},
  \bibinfo{year}{2014}.
\newblock \bibinfo{title}{Auto-encoding variational {Bayes}}, in:
  \bibinfo{booktitle}{Proc. ICLR'14}.
\bibitem[{Kosinski(1999)}]{key-30}
\bibinfo{author}{Kosinski, A.}, \bibinfo{year}{1999}.
\newblock \bibinfo{title}{A procedure for the detection of multivariate
  outliers}.
\newblock \bibinfo{journal}{Computational Statistics and Data Analysis}
  \bibinfo{volume}{29}, \bibinfo{pages}{145--161}.
\bibitem[{Le et~al.(2013)Le, Sarl\'os and Smola}]{fastfood}
\bibinfo{author}{Le, Q.}, \bibinfo{author}{Sarl\'os, T.},
  \bibinfo{author}{Smola, A.}, \bibinfo{year}{2013}.
\newblock \bibinfo{title}{Fastfood --- approximating kernel expansions in
  loglinear time}, in: \bibinfo{booktitle}{Proc. ICML}.
\bibitem[{LeCun et~al.(2015)LeCun, Bengio and Hinton}]{lecun}
\bibinfo{author}{LeCun, Y.}, \bibinfo{author}{Bengio, Y.},
  \bibinfo{author}{Hinton, G.}, \bibinfo{year}{2015}.
\newblock \bibinfo{title}{Deep learning}.
\newblock \bibinfo{journal}{Nature} \bibinfo{volume}{512},
  \bibinfo{pages}{436--444}.
\bibitem[{Liu and Rubin(1995)}]{liu}
\bibinfo{author}{Liu, C.}, \bibinfo{author}{Rubin, D.}, \bibinfo{year}{1995}.
\newblock \bibinfo{title}{{ML} estimation of the $t$ distribution using {EM}
  and its extensions, {ECM} and {ECME}}.
\newblock \bibinfo{journal}{Statistica Sinica} \bibinfo{volume}{5},
  \bibinfo{pages}{19--39}.
\bibitem[{Loosli et~al.(2007)Loosli, Canu and Bottou}]{mnist8m}
\bibinfo{author}{Loosli, G.}, \bibinfo{author}{Canu, S.},
  \bibinfo{author}{Bottou, L.}, \bibinfo{year}{2007}.
\newblock \bibinfo{title}{Training invariant support vector machines using
  selective sampling}, in: \bibinfo{editor}{Bottou, L.},
  \bibinfo{editor}{Chapelle, O.}, \bibinfo{editor}{DeCoste, D.},
  \bibinfo{editor}{Weston, J.} (Eds.), \bibinfo{booktitle}{Large Scale Kernel
  Machines}. \bibinfo{publisher}{MIT Press}, \bibinfo{address}{Cambridge, MA},
  pp. \bibinfo{pages}{301--320}.
\bibitem[{McLachlan and Peel(2000)}]{key-19}
\bibinfo{author}{McLachlan, G.}, \bibinfo{author}{Peel, D.},
  \bibinfo{year}{2000}.
\newblock \bibinfo{title}{Finite Mixture Models}.
\newblock \bibinfo{publisher}{Wiley Series in Probability and Statistics},
  \bibinfo{address}{New York}.
\bibitem[{Nair and Hinton(2010)}]{relu}
\bibinfo{author}{Nair, V.}, \bibinfo{author}{Hinton, G.}, \bibinfo{year}{2010}.
\newblock \bibinfo{title}{Rectified linear units improve restricted {Boltzmann}
  machines}, in: \bibinfo{booktitle}{Proc. ICML}.
\bibitem[{Naudts(2002)}]{naudts}
\bibinfo{author}{Naudts, J.}, \bibinfo{year}{2002}.
\newblock \bibinfo{title}{Deformed exponentials and logarithms in generalized
  thermostatistics}.
\newblock \bibinfo{journal}{Physica A} \bibinfo{volume}{316},
  \bibinfo{pages}{323-- 334}.
\bibitem[{Naudts(2004a)}]{naudts3}
\bibinfo{author}{Naudts, J.}, \bibinfo{year}{2004}a.
\newblock \bibinfo{title}{Estimators, escort proabilities, and
  $\phi$-exponential families in statistical physics}.
\newblock \bibinfo{journal}{Journal of Inequalities in Pure and Applied
  Mathematics} \bibinfo{volume}{5}.
\bibitem[{Naudts(2004b)}]{naudts2}
\bibinfo{author}{Naudts, J.}, \bibinfo{year}{2004}b.
\newblock \bibinfo{title}{Generalized thermostatistics and mean-field theory}.
\newblock \bibinfo{journal}{Physica A} \bibinfo{volume}{332},
  \bibinfo{pages}{279--300}.
\bibitem[{Neal(1995)}]{neal}
\bibinfo{author}{Neal, R.M.}, \bibinfo{year}{1995}.
\newblock \bibinfo{title}{Bayesian learning for neural networks}.
\newblock Ph.D. thesis. University of Toronto.
\bibitem[{Partaourides and Chatzis(2017)}]{asydgm}
\bibinfo{author}{Partaourides, H.}, \bibinfo{author}{Chatzis, S.P.},
  \bibinfo{year}{2017}.
\newblock \bibinfo{title}{Asymmetric deep generative models}.
\newblock \bibinfo{journal}{Neurocomputing} \bibinfo{volume}{241},
  \bibinfo{pages}{90--96}.
\bibitem[{Pedregosa et~al.(2011)Pedregosa, Varoquaux, Gramfort, Michel,
  Thirion, Grisel, Blondel, Prettenhofer, Weiss, Dubourg, Vanderplas, Passos,
  Cournapeau, Brucher, Perrot and Duchesnay}]{scikit-learn}
\bibinfo{author}{Pedregosa, F.}, \bibinfo{author}{Varoquaux, G.},
  \bibinfo{author}{Gramfort, A.}, \bibinfo{author}{Michel, V.},
  \bibinfo{author}{Thirion, B.}, \bibinfo{author}{Grisel, O.},
  \bibinfo{author}{Blondel, M.}, \bibinfo{author}{Prettenhofer, P.},
  \bibinfo{author}{Weiss, R.}, \bibinfo{author}{Dubourg, V.},
  \bibinfo{author}{Vanderplas, J.}, \bibinfo{author}{Passos, A.},
  \bibinfo{author}{Cournapeau, D.}, \bibinfo{author}{Brucher, M.},
  \bibinfo{author}{Perrot, M.}, \bibinfo{author}{Duchesnay, E.},
  \bibinfo{year}{2011}.
\newblock \bibinfo{title}{Scikit-learn: Machine learning in {P}ython}.
\newblock \bibinfo{journal}{Journal of Machine Learning Research}
  \bibinfo{volume}{12}, \bibinfo{pages}{2825--2830}.
\bibitem[{Platanios and Chatzis(2014)}]{gpmch}
\bibinfo{author}{Platanios, E.A.}, \bibinfo{author}{Chatzis, S.P.},
  \bibinfo{year}{2014}.
\newblock \bibinfo{title}{Gaussian process-mixture conditional
  heteroscedasticity}.
\newblock \bibinfo{journal}{IEEE Transactions on Pattern Analysis and Machine
  Intelligence} \bibinfo{volume}{36}, \bibinfo{pages}{888--900}.
\bibitem[{Rahimi and Recht(2009)}]{kitchen}
\bibinfo{author}{Rahimi, A.}, \bibinfo{author}{Recht, B.},
  \bibinfo{year}{2009}.
\newblock \bibinfo{title}{{Weighted Sums of Random Kitchen Sinks: Replacing
  minimization with randomization in learning}}, in: \bibinfo{booktitle}{Proc.
  NIPS}.
\bibitem[{Rudin(2011)}]{bochner}
\bibinfo{author}{Rudin, W.}, \bibinfo{year}{2011}.
\newblock \bibinfo{title}{Fourier Analysis on Groups}.
\newblock \bibinfo{publisher}{John Wiley \& Sons}.
\bibitem[{Schapire(2003)}]{adaboost}
\bibinfo{author}{Schapire, R.E.}, \bibinfo{year}{2003}.
\newblock \bibinfo{title}{The boosting approach to machine learning: An
  overview}.
\newblock Nonlinear Estimation and Classification,
  \bibinfo{publisher}{Springer}.
\bibitem[{Silver et~al.(2016)}]{silver}
\bibinfo{author}{Silver, D.}, et~al., \bibinfo{year}{2016}.
\newblock \bibinfo{title}{Mastering the game of go with deep neural networks
  and tree search}.
\newblock \bibinfo{journal}{Nature} \bibinfo{volume}{529},
  \bibinfo{pages}{484-- 489}.
\bibitem[{Sousa and Tsallis(1994)}]{tsalis2}
\bibinfo{author}{Sousa, A.}, \bibinfo{author}{Tsallis, C.},
  \bibinfo{year}{1994}.
\newblock \bibinfo{title}{Student's $t$- and $r$-distributions: Unified
  derivation from an entropic variational principle}.
\newblock \bibinfo{journal}{Physica A} \bibinfo{volume}{236},
  \bibinfo{pages}{52--57}.
\bibitem[{Srivastava et~al.(2014)Srivastava, Hinton, Krizhevsky, Sutskever and
  Salakhutdinov}]{dropout}
\bibinfo{author}{Srivastava, N.}, \bibinfo{author}{Hinton, G.},
  \bibinfo{author}{Krizhevsky, A.}, \bibinfo{author}{Sutskever, I.},
  \bibinfo{author}{Salakhutdinov, R.}, \bibinfo{year}{2014}.
\newblock \bibinfo{title}{Dropout: a simple way to prevent neural networks from
  overfitting}.
\newblock \bibinfo{journal}{Journal of Machine Learning Research}
  \bibinfo{volume}{15}, \bibinfo{pages}{1929--1958}.
\bibitem[{Sutskever et~al.(2014)Sutskever, Vinyals and Le}]{sutskever}
\bibinfo{author}{Sutskever, I.}, \bibinfo{author}{Vinyals, O.},
  \bibinfo{author}{Le, Q.V.}, \bibinfo{year}{2014}.
\newblock \bibinfo{title}{Sequence to sequence learning with neural networks},
  in: \bibinfo{booktitle}{Proc. NIPS}.
\bibitem[{Svens\'en and Bishop(2005a)}]{vbsmmBishop}
\bibinfo{author}{Svens\'en, M.}, \bibinfo{author}{Bishop, C.M.},
  \bibinfo{year}{2005}a.
\newblock \bibinfo{title}{Robust {Bayesian} mixture modelling}.
\newblock \bibinfo{journal}{Neurocomputing} \bibinfo{volume}{64},
  \bibinfo{pages}{235--252}.
\bibitem[{Svens\'en and Bishop(2005b)}]{vbsmm2}
\bibinfo{author}{Svens\'en, M.}, \bibinfo{author}{Bishop, C.M.},
  \bibinfo{year}{2005}b.
\newblock \bibinfo{title}{Robust {Bayesian} mixture modelling}.
\newblock \bibinfo{journal}{Neurocomputing} \bibinfo{volume}{64},
  \bibinfo{pages}{235--252}.
\bibitem[{Tsallis(1998)}]{tsalis}
\bibinfo{author}{Tsallis, C.}, \bibinfo{year}{1998}.
\newblock \bibinfo{title}{Possible generalization of {Boltzmann-Gibbs}
  statistics}.
\newblock \bibinfo{journal}{J. Stat. Phys.} \bibinfo{volume}{52},
  \bibinfo{pages}{479--487}.
\bibitem[{Tsallis et~al.(1998)Tsallis, Mendes and Plastino}]{tsalis3}
\bibinfo{author}{Tsallis, C.}, \bibinfo{author}{Mendes, R.S.},
  \bibinfo{author}{Plastino, A.R.}, \bibinfo{year}{1998}.
\newblock \bibinfo{title}{The role of constraints within generalized
  nonextensive statistics.}
\newblock \bibinfo{journal}{Physica A} \bibinfo{volume}{261},
  \bibinfo{pages}{534--554}.
\bibitem[{Vapnik(1998)}]{svm-book}
\bibinfo{author}{Vapnik, V.N.}, \bibinfo{year}{1998}.
\newblock \bibinfo{title}{Statistical Learning Theory}.
\newblock \bibinfo{publisher}{Wiley}, \bibinfo{address}{New York}.
\bibitem[{Wainwright and Jordan(2008)}]{entropy}
\bibinfo{author}{Wainwright, M.J.}, \bibinfo{author}{Jordan, M.I.},
  \bibinfo{year}{2008}.
\newblock \bibinfo{title}{Graphical models, exponential families, and
  variational inference}.
\newblock \bibinfo{journal}{Foundations and Trends in Machine Learning}
  \bibinfo{volume}{1}, \bibinfo{pages}{1--305}.
\bibitem[{Yang et~al.(2015)Yang, Smola, Song and Wilson}]{alacarte}
\bibinfo{author}{Yang, Z.}, \bibinfo{author}{Smola, A.J.},
  \bibinfo{author}{Song, L.}, \bibinfo{author}{Wilson, A.G.},
  \bibinfo{year}{2015}.
\newblock \bibinfo{title}{{\'A la Carte --- Learning Fast Kernels}}, in:
  \bibinfo{booktitle}{Proc. AISTATS}.

\end{thebibliography}

\end{document}